\pdfoutput=1

\documentclass[11pt]{article}

\usepackage[final]{acl}

\usepackage{times}
\usepackage{latexsym}

\usepackage[T1]{fontenc}

\usepackage[utf8]{inputenc}

\usepackage{microtype}

\usepackage{inconsolata}

\usepackage{graphicx}
\usepackage{multirow} 

\title{Change My Frame: Reframing in the Wild in r/ChangeMyView}

\author{Arturo Mart\'inez Peguero \and Taro Watanabe \\
        NAIST \\
        \texttt{arturoComputes@gmail.com}, \texttt{taro@is.naist.jp}}

\begin{document}
\maketitle
\begin{abstract}
    Recent work in reframing, within the scope of text style transfer, has so far made use of out-of-context, task-prompted utterances in order to produce neutralizing or optimistic reframes.
    Our work aims to generalize reframing based on the subreddit r/ChangeMyView (CMV).
    We build a dataset that leverages CMV's community's interactions and conventions to identify high-value, community-recognized utterances that produce changes of perspective.
    With this data, we widen the scope of the direction of reframing since the changes in perspective do not only occur in neutral or positive directions.
    We fine tune transformer-based models, make use of a modern LLM to refine our dataset, and explore challenges in the dataset creation and evaluation around this type of reframing.
\end{abstract}

\section{Introduction}\label{sec:introduction}

    \textit{Reframing} is a text style transfer technique that alters the frame, perspective, or focus of an utterance, aiming to highlight different aspects of the content while preserving its original meaning.
    A reframed sentence is compatible with the original, yet it shifts the perspective to emphasize different features of the situation, making specific contextual aspects more salient. 
    Reframing does not contradict the core content and meaning of the original sentence; rather, it often introduces new content derived from a broader context, providing a distinct perspective or focus.
    A typical
    example of reframing involves a glass being half \textit{full} or half \textit{empty}, respectively denoting optimism or pessimism.

    The perspective that an utterance implicitly carries can dramatically change the impact it has on a listener.
    The way we frame our communications has uses as benign as therapy, and as harmful as manipulative propaganda.
    
    Recent work has shown the possibility of crowdsourcing reframing datasets and
    performing \textit{positive reframing}, where a sentence can be rephrased with a different and positive perspective while preserving its meaning.
    In this work,
    we avoid data that is obtained in unnatural experiment-task-based contexts 
    both in order to make efficient use of publicly available forums and to reduce costs of hiring crowdsource workers or experts.
    Our \texttt{(original,reframing)} pairs are not prompted artificially but rather observed in the more naturally-occuring environment of one of the most popular public discussion forums in the world: Reddit.

    In particular, we look at the subreddit r/ChangeMyView (CMV) 
    since it 
    self-describes as a forum where different perspectives on 
    a single issue are shared.

    \begin{figure}
        \centering
        \includegraphics[scale=0.125]{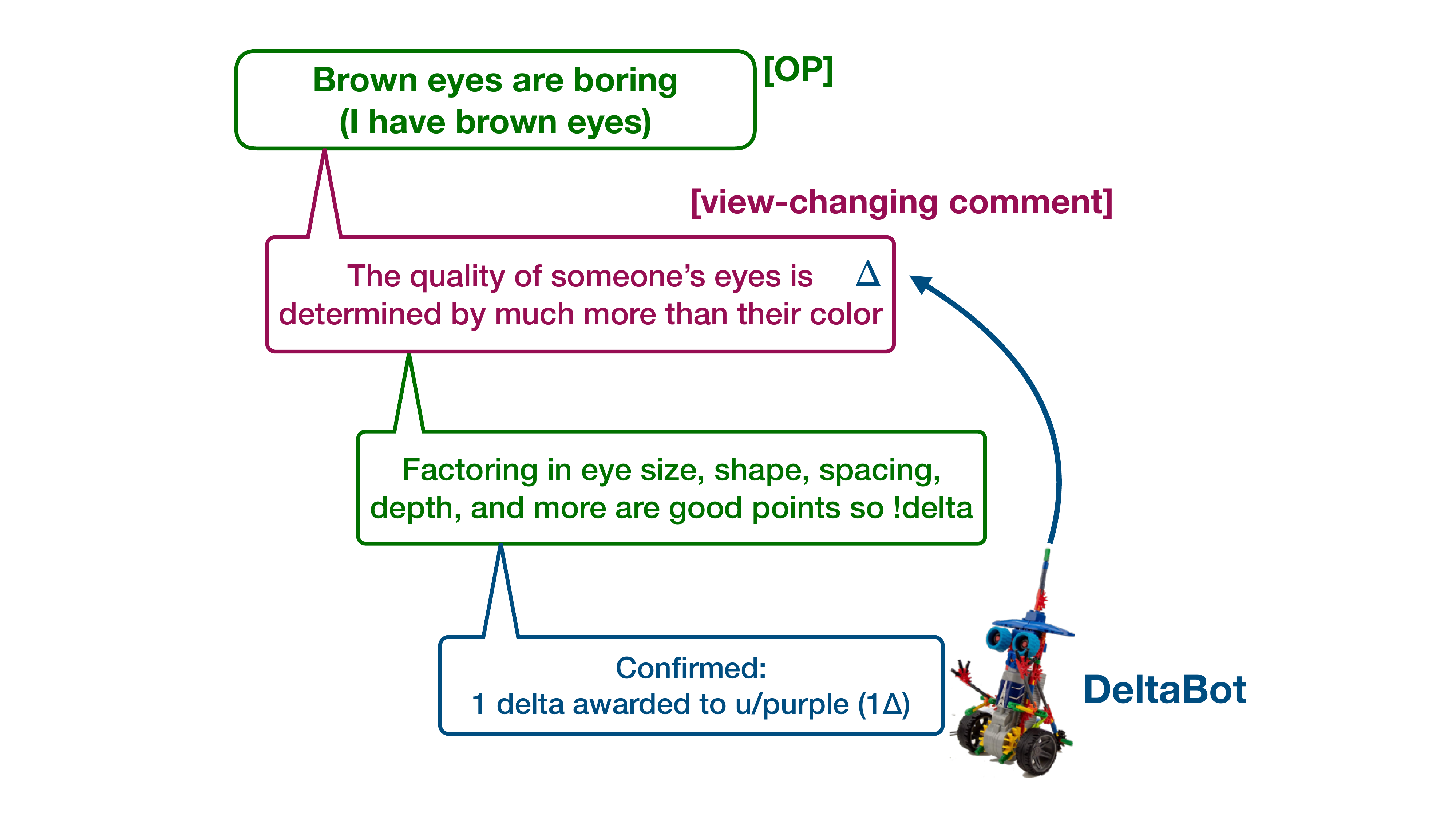}
        \caption{Data collection mechanism described in section \ref{subsec:leveragingdelta}. We look for posts' direct replies that change views and that are recognized as such by the original post (OP) author.}
        \label{fig:direct-reply-delta}
    \end{figure}

    Our dataset currently consists of
    32,306~\texttt{(post,comment)}
    pairs identified by CMV community members to be of high value and to have changed a view,
    as shown on Figure~\ref{fig:direct-reply-delta}.
    Fine-tuning on our dataset has limited but increasing success as 
    the dataset is pared down.
    Finally, challenges with proper reframing evaluation remain.

\section{Previous Work}\label{sec:previous-work}
%
    
    Some recent studies have explicitly addressed positive reframing.
    \citet{ziemsli2022reframing} constructed a parallel dataset of original sentences reflecting negative, stress-related tweets, along with crowdsourced, manually crafted novel reframings.
    With these dataset and psychological strategies, they used transformer-based models and succeeded in performing \textit{positive} reframings.

    More recently, \citet{sharma2023cognitive} created a mental-health-expert-defined framework of linguistic-attributes that contribute to reframing.
    Experts generated reframes from a database of negative thoughts and the framework helped automatically evaluate reframing attributes.
    Additionally, \citet{maddela2023training} asked crowdsourced to generate original sentences, to label such sentences and to reframe them.
    Both studies built reframing models from their data.
    
    Since the studies mentioned
    addressed mostly \textit{positive reframing} and used crowdsourcing as their main source of reframes, 
    that has opened the door to other kinds of reframings.

\section{Dataset Building}

    \subsection{r/ChangeMyView}

        The subreddit r/ChangeMyView (CMV) is an online social forum that hosts a community around the goal of changing each others' views on particular topics.
    
        The basic mechanics of CMV consist of an \textit{original post} (OP) author submitting a concept about which they hold an original view.
        When they post in CMV, they are asking the community to change their view.
        Other CMV members then reply to the OP presenting different perspectives on the issue.
        Any submission that is not the OP is referred to as a \textit{comment}.
        
        Starting from the OP, comments themselves can have their own subcomments and so on recursively.
        Indeed, any particular comment can be embedded in an arbitrary heavily-nested depth below the OP.
    
        When a particular contribution is deemed to be valuable to the Reddit community, recognitions such as upvotes, gold and flair are awarded.
        On top of these, the CMV community also assigns \textit{deltas}.
        Borrowing from the mathematical use of the greek letter $\Delta$ (uppercase delta) to indicate change, CMV members award deltas to recognize comments that have changed their perspective relative to the OP.
        This is done through the \textit{DeltaBot}, a bot that monitors the CMV subreddit and awards deltas automatically when a user indicates their desire to assign a delta to a particular comment.

    \subsection{Leveraging the delta system}\label{subsec:leveragingdelta}

        Given the
        CMV specifics discussed above,
        we build a
        dataset
        on comments that have changed an
        OP author's
        particular perspective.
        
        We filter out moderation posts, as well as \textit{[deleted]}, \textit{[removed]} and empty posts or comments, since they are not helpful for reframing purposes,
        We 
        also only
        consider 
        comments that have been awarded deltas.
        We exploit the delta-awarding mechanics
        and DeltaBot comment structure
        to obtain 
        \texttt{(post,comment)} 
        pairs consisting of (1) an OP, and (2) a delta-awarded comment.
        Given moderator-enforced rules around post and comment lengths,
        we filter out any pairs where the post is shorter than 500 characters
        and where the comment is shorter than 100 characters.
        Further, we consider only posts 
        that explicitly indicate
        that the OP author
        themselves
        has awarded a delta.
        To tighten the constraints,
        we limit consideration
        of 
        \texttt{(post,comment)}
        pairs
        to those where the delta-awarded comment
        is a direct reply
        (i.e. not nested deep in the replies)
        to the OP.
        Additionally, only delta-awarding comments by the OP author are taken into account.
        This is illustrated in Figure \ref{fig:direct-reply-delta}.

    \subsection{r/ChangeMyView reframing data}

        The full initial dataset consisted of
        around 
        255,287
        posts and 
        11,461,626
        comments
        from CMV
        ranging from
        early 2012
        to the end of 2022.
        Post and comment bodies as well as metadata was included in the
        dataset.
        Since flair
        labels
        indicating that an OP author awarded deltas
        only appear from 2015 onward,
        only 2015-2022 data was used.

        Given the restrictions described 
        in the previous section,
        we built a dataset of
        32,306
        \texttt{(post,comment)}
        pairs
        where the original author
        awarded a delta to a direct reply to their OP.

\begin{table*}
    \centering
    \begin{tabular}{ll|cccccc}
        \hline
        \textbf{Dataset}                                & \textbf{Model}    & \textbf{R-1}  & \textbf{R-2}  & \textbf{R-L}  & \textbf{BLEU} & \textbf{BERTScore} \\ 
        \hline
        \multirow{4}{*}{\citet{ziemsli2022reframing}}     & \textbf{T5}       & 27.4          & 9.8           & 23.8          & 8.7           &   88.7 \\ 
                                                        & \textbf{BART}     & 27.7          & 10.8          & 24.3          & 10.3          &   89.3 \\
                                                        & \textbf{GPT}      & 13.3          & 1.8           & 11.3          & 1.1           &   86.4 \\
                                                        & \textbf{GPT-2 NoPretrain}  & 13.2 & 1.3           & 11.4          & 0.66          &   89.6\\
        \hline
        \multirow{2}{*}{LLM-ext.pair, 3.6K shortest}      & \textbf{T5}       & 14.7         & 2.51           & 11.2          & 0.1             &   85.0  \\
                                                        & \textbf{BART}     & 14.7         & 2.73           & 11.1          & 0.1             &   85.3 \\ 
        \hline
        \multirow{2}{*}{original pair, 2K random}         & \textbf{T5}       & 14.2         & 2.47          & 10.7         & 0.2           &   84.1 \\
                                                        & \textbf{BART}     & 13.7         & 1.99          & 10.2         & 0.1           &   84.7 \\ 
        \hline
        \multirow{2}{*}{original pair, 4K random}         & \textbf{T5}       & 12.9         & 2.08          & 10.0          & 0.0             &   84  \\
                                                        & \textbf{BART}     & 12.8         & 2.1          & 9.38          & 0.0             &   84.4 \\ 
        
        \hline
    \end{tabular}
    \caption{Comparison of results between \citet{ziemsli2022reframing} and our work. ROUGE-1 (R-1), ROUGE-2 (R-2), ROUGE-L (R-L), BLEU \& BERTScore (BScore) are shown. The best performance for our model is achieved with fewer, shorter utterances or by distilling the existing utterances to their main reframing components. Doing this, performance approaches but does not yet match \citet{ziemsli2022reframing}'s. A full table of results is included in Appendix~\ref{sec:appendix1-results}.}
    \label{tab:results01}
\end{table*}

\begin{table}
    \centering
    \begin{tabular}{ll|cccccc}
        \hline
        \textbf{FineT-eval data}            & \textbf{Model}    & \textbf{R-1}  & \textbf{R-2}  & \textbf{R-L} \\ 
        \hline
        \multirow{2}{*}{Ziems et al.'s-ours}     & \textbf{T5}       & 8.16          & 1.1           & 6.16 \\
                                    & \textbf{BART}     & 8.64          & 1.27          & 6.47 \\
        \hline
        \multirow{2}{*}{ours-Ziems et al.'s}     & \textbf{T5}       & 14.0          & 2.32          & 10.5 \\
                                    & \textbf{BART}     & 13.7          & 1.97          & 10.2 \\
        \hline
    \end{tabular}
    \caption{Best results from fine-tuning with \citet{ziemsli2022reframing}'s data while evaluating with one of our dataset subsets, and viceversa. BLEU scores are all 0 or near-0 and are omitted.}
    \label{tab:results02}
\end{table}

\section{Experiments \& Evaluation}

    For our experiments,
    for ease of comparison with 
    previous 
    work in reframing,
    we 
    follow
    the experiment design by \citet{ziemsli2022reframing}
    using our 
    dataset 
    to fine-tune
    the
    encoder-decoder 
    models 
    BART \citep{lewis2020bart}
    and 
    T5~\citep{raffel2019t5}
    with greedy decoding.
    We 
    assign 80\% of the
    \texttt{(post,comment)}
    pairs
    to the training split,
    10\% to the development split
    and 10\% to the test split.
    In order 
    to 
    overcome
    memory limitations,
    we run multiple experiments with subsets of our full dataset.
    For evaluation,
    we use overlap-based similarity measures
    BLEU~\citep{papineni2002bleu},
    ROUGE~\citep{lin2004rouge}
    as well as semantic-similarity measure
    BERTScore.\citep{zhang2019bertscore}

    We also use GPT-4 \cite{openai2024gpt4} to trim down the
    \texttt{(post,comment)} pairs to the most relevant
    \texttt{(original text,reframing)} pairs and fine-tune with a subset of
    3,624 such datapoints.

\section{Results \& Discussion}

    As shown on Table \ref{tab:results01},
    our work 
    so far 
    has only managed to achieve performance similar to that
    of GPT or no-pretrain-GPT-2 from \citet{ziemsli2022reframing}
    on overlap-similarity measures.
    BERTScore increased as we took away data,
    either 
    with
    fewer sentences
    or by trimming 
    data, assisted by GPT-4.
    
    Additional cross-dataset experiments were conducted by (A) fine-tuning on \citet{ziemsli2022reframing} data and evaluating on our test data, and (B) fine-tuning on our data and evaluating on test data from \citet{ziemsli2022reframing}.
    These results are shown on Table \ref{tab:results02}
    and suggest
    that our dataset 
    affects performance negatively,
    causing suboptimal fine-tuning
    and setting up a difficult testing environment.

    Fine-tuning on our data and testing on \citet{ziemsli2022reframing} data seemed to yield better results
    on overlap-similarity measures than the other way around, indicating potentially more 
    universality and tolerance for diversity 
    from our data.

    These results suggests that
    our
    dataset 
    still contains too much context around
    the original exposition of a particular point of view
    and around
    the 
    substring
    of the comment that contains the reframing.
    The experiments with GPT-4-aided pair trimming seem to confirm this,
    obtaining the highest BERTscore and overlap-based performances
    once unnecessary text had been reduced or eliminated.

\section{Conclusion and future work}

    Our results point to the need to pare down our data, either manually or with the help of LLMs.
    This will also open the doors to few-shot learning, explored in some of the studies mentioned in Section \ref{sec:previous-work}.
    
    Reframing, which demands meaning preservation but keeps open the possibility of new content,
    is not completely adequately served by surface-form similarity evaluations.
    While BERTScore provides some degree of richer semantic tolerance and diversity of formulation of an idea,
    measures that automatically assess different parts of a reframe, such as the linguistic attribute framework suggested in \citet{sharma2023cognitive}, can contribute to a fuller evaluation.

    Finally, extending this work to other languages, especially Spanish, has been and remains a focus of high interest to us.

\bibliography{acl_latex}

\appendix

\section{Appendix A: Results from more experiments}
\label{sec:appendix1-results}

Table \ref{tab:results03} below shows the results for all the experiments done, not shown earlier due to length requirements. Despite having access to 32,306 \texttt{(post,comment)} pairs, memory limitations made it difficult to run any experiments beyond 8,000 pairs for random pairs and 10,000 for shortest pairs.

\begin{table*}
    \centering
    \begin{tabular}{ll|cccccc}
        \hline
        \textbf{Dataset}                                & \textbf{Model}    & \textbf{R-1}  & \textbf{R-2}  & \textbf{R-L}  & \textbf{BLEU} & \textbf{BERTScore} \\ 
        \hline
        \multirow{4}{*}{\citet{ziemsli2022reframing}}   & \textbf{T5}       & 27.4          & 9.8           & 23.8          & 8.7           &   88.7 \\ 
                                                        & \textbf{BART}     & 27.7          & 10.8          & 24.3          & 10.3          &   89.3 \\ 
                                                        & \textbf{GPT}      & 13.3          & 1.8           & 11.3          & 1.1           &   86.4 \\
                                                        & \textbf{GPT-2 NoPretrain}  & 13.2 & 1.3           & 11.4          & 0.66          &   89.6\\
        \hline
        \multirow{2}{*}{LLM-ext.pair, 3.6K shortest}    & \textbf{T5}       & 14.7         & 2.51           & 11.2          & 0.1             &   85.0  \\ 
                                                        & \textbf{BART}     & 14.7         & 2.73           & 11.1          & 0.1             &   85.3 \\  
        \hline
        \multirow{2}{*}{original pair, 2K random}       & \textbf{T5}       & 14.2         & 2.47          & 10.7         & 0.2           &   84.1 \\
                                                        & \textbf{BART}     & 13.7         & 1.99          & 10.2         & 0.1           &   84.7 \\
        \hline
        \multirow{2}{*}{original pair, 2K shortest}     & \textbf{T5}       & 14.1         & 2.45          & 10.7         & 0.2           &   84.1 \\ 
                                                        & \textbf{BART}     & 13.7          & 2.0          & 10.2           & 0.1           &   84.7 \\ 
        \hline
        \multirow{2}{*}{original pair, 4K random}       & \textbf{T5}       & 12.9         & 2.08          & 10.0          & 0.0             &   84.0  \\ 
                                                        & \textbf{BART}     & 12.8         & 2.1          & 9.38          & 0.0             &   84.4 \\ 
        \hline
        \multirow{2}{*}{original pair, 4K shortest}     & \textbf{T5}       & 12.9         & 2.1           & 10.0          & 0.0             &   84.0  \\ 
                                                        & \textbf{BART}     & 12.8          & 2.11          & 9.4          & 0.0             &   84.4 \\ 
        \hline
        \multirow{2}{*}{original pair, 6K random}       & \textbf{T5}       & 12.9         & 1.81          & 9.57          & 0.0             &   84.0  \\ 
                                                        & \textbf{BART}     & 12.6          & 2.12          & 9.5           & 0.0             &   84.2 \\ 
        \hline
        \multirow{2}{*}{original pair, 6K shortest}     & \textbf{T5}       & 12.8         & 1.75          & 9.54          & 0.0             &   83.9 \\ 
                                                        & \textbf{BART}     & 12.6          & 2.12          & 9.5           & 0.0             &   84.2 \\ 
        \hline
        \multirow{2}{*}{original pair, 8K random}       & \textbf{T5}       & 11.9         & 1.86          & 9.02          & 0.0             &   83.9 \\ 
                                                        & \textbf{BART}     & 12.2          & 2.18          & 9.09          & 0.0             &   84.1 \\
        \hline
        \multirow{2}{*}{original pair, 8K shortest}     & \textbf{T5}       & 11.9         & 1.86          & 9.03          & 0.0             &   83.9 \\ 
                                                        & \textbf{BART}     & 12.2          & 2.19          & 9.1          & 0.0             &   84.1 \\
        \hline
        \multirow{2}{*}{original pair, 10K shortest}    & \textbf{T5}       & 12.2         & 2.07          & 9.2          & 0.0             &   83.9 \\ 
                                                        & \textbf{BART}     & 11.5          & 2.05          & 8.54          & 0.0             &   84.1 \\
        \hline
    \end{tabular}
    \caption{Comparison of results between \citet{ziemsli2022reframing} and our work. ROUGE-1 (R-1), ROUGE-1 (R-2), ROUGE-L (R-L), BLEU \& BERTScore (BScore) are shown. The best performance for our model is achieved with fewer, shorter utterances or by distilling the existing utterances to their main reframing components. Doing this, performance approaches but does not yet match \citet{ziemsli2022reframing}'s.}
    \label{tab:results03}
\end{table*}

\end{document}